\documentclass[twocolumn,10pt]{article}

\usepackage{times}
\usepackage{soul}
\usepackage{url}
\usepackage[utf8]{inputenc}
\usepackage[small]{caption}
\usepackage{graphicx}
\usepackage{amsmath}
\usepackage{amssymb}
\usepackage{amsfonts}
\usepackage{amsthm}
\usepackage{booktabs}
\usepackage{algorithm}
\usepackage{algorithmic}
\usepackage{xcolor}
\usepackage{pgf}
\usepackage{tikz,pgfplots}
\pgfplotsset{compat=1.14}

\usepackage{graphicx}

\usepackage{multirow}

\usepackage[normalem]{ulem}
\usepackage{pifont}

\newcommand{\unarycon}{$\mathbf{U_c}$}
\newcommand{\unaryex}{$\mathbf{U_d}$}
\newcommand{\binary}{$\mathbf{B}$}

\title{Reinforced Anytime Bottom Up Rule Learning for\\ Knowledge Graph Completion}

\author{%
Christian Meilicke$^1$,
Melisachew Wudage Chekol$^2$, \\
Manuel Fink$^1$,
Heiner Stuckenschmidt$^1$  \\
$^1$Research Group Data and Web Science, University Mannheim\\
$^2$Utrecht University\\
\{christian,manuel,heiner\}@informatik-uni-mannheim.de, m.w.chekol@uu.nl}

\date{ }

\begin{document}

\maketitle

\begin{abstract}
Most of today’s work on knowledge graph completion is concerned with sub-symbolic approaches that focus on the concept of embedding a given graph in a low dimensional vector space. Against this trend, we propose an approach called AnyBURL that is rooted in the symbolic space. Its core algorithm is based on sampling paths, which are generalized into Horn rules. Previously published results show that the prediction quality of AnyBURL is on the same level as current state of the art with the additional benefit of offering an explanation for the predicted fact. In this paper, we are concerned with two extensions of AnyBURL. Firstly, we change AnyBURL’s interpretation of rules from $\Theta$-subsumption into $\Theta$-subsumption under Object Identity. Secondly, we introduce reinforcement learning to better guide the sampling process. We found out that reinforcement learning helps finding more valuable rules earlier in the search process. We measure the impact of both extensions and compare the resulting approach with current state of the art approaches. Our results show that AnyBURL outperforms most sub-symbolic methods.
\end{abstract}

\section{Introduction}
\label{sec:intro}

%


Knowledge graphs (KGs) are widely employed in various domains. Examples are FreeBase~\cite{bollacker2008freebase}, DBPedia~\cite{auer2007dbpedia}, YAGO~\cite{suchanek2007yago}, Google Knowledge Graph and Microsoft Satori. These massive KGs can contain up to millions of entities and billions of facts. As pointed out in~\cite{dong2014knowledge}, knowledge graphs are often incomplete. The task to construct missing triples using the vocabulary already used in the graph is known as \emph{knowledge graph completion} or \emph{link prediction}. This task can be solved with the additional help of external resources (e.g., text in web-pages) or by inferring new triples solely from the triples in a given knowledge graph. We are concerned with the latter problem.

An approach that does not use external information must rely on the statistics, patterns, distributions or any other kind of regularity that can be found in the given knowledge graph. An intuitive choice for solving such a task is to learn and apply an explicit, symbolic representation of these patterns. While there is long history of approaches that are concerned with learning symbolic representations, such as inductive logic programming~\cite{muggleton1994inductive} and relational association rule mining~\cite{dehaspe2001discovery}, today's research is following a different paradigm. The vast majority of methods that are developed nowadays learn a low dimensional, sub-symbolic representation of a given knowledge graph. Inspired by early models such as RESCAL~\cite{nickel2011three} and TransE~\cite{transe2013}, a large number of new models have been developed within the last decade. As a result, symbolic approaches are underrepresented in knowledge graph completion research. 




We have developed a symbolic approach~\cite{meilicke2019anyburl} with a language bias that mines especially those rules that might be relevant for the task at hand. We called our approach AnyBURL (Anytime Bottom-up Rule Learning) due to its anytime behaviour and fact that it is based on a sampling component that generalizes paths into rules. Our results as well as the results reported in an independent evaluation of the current state of the art~\cite{rossi2020knowledge} revealed that AnyBURL is not just a symbolic baseline, but performs on the same level as the best models proposed in the last five years. In this paper, we further improve AnyBURL and report about the impact of two extensions.


With this paper we give for the first time an elaborate description of AnyBURL. Aside from the original algorithm, we describe our extensions and improvements and report about comprehensive experiments. In particular, the paper contains the following contributions.
\begin{itemize}
	\item We take up the concept of Object Identity~\cite{semeraro1994avoiding} and report about experiments that illustrate its benefits w.r.t knowledge graph completion. Our results show that it prevents learning  a large number of quasi-redundant rules with misleading confidence scores.
	\item We introduce reinforcement learning to guide the search during sampling paths. We argue that reinforcement learning is more robust, allows to leverage the specifics of a given knowledge graph, and is less affected by choosing a wrong parameter setting.
\end{itemize}
The results of our experiments show that the improved version of AnyBURL is one of the best approaches available for the knowledge graph completion task.

\section{Bottom Up Rule Learning}
\label{sec:learn}

We first introduce the type of rules that can be learned by AnyBURL before we describe how we create these rules from sampling paths. Parts of this were already presented in a different form in~\cite{meilicke2019anyburl}. Then we explain the concept of Object Identity that was introduced in~\cite{semeraro1994avoiding}, and argue why it is important for our use case. Object Identity was partially implemented in the previous version of AnyBURL without understanding its importance.

\subsection{Language Bias}
\label{sub:learn-language}

We distinguish in the following between three types of rules that we call binary rules (\binary), unary rules ending with a dangling atom (\unaryex) and unary rules ending with an atom that includes a constant (\unarycon)\footnote{In~\cite{meilicke2019anyburl} we called binary rules cyclic rules and unary rules acyclic rules. This convention was slightly confusing, because a unary rule can also be sampled from a cyclic path.}.
\begin{align*}
\text{\binary}   && h(A_0,A_n)  \leftarrow \bigwedge_{i=1}^{n} b_i(A_{i-1}, A_i) \\
\text{\unaryex}  && h(A_0,c)    \leftarrow \bigwedge_{i=1}^{n} b_i(A_{i-1}, A_i) \\
\text{\unarycon} && h(A_0,c)    \leftarrow \left(\bigwedge_{i=1}^{n-1} b_i(A_{i-1}, A_i)\right) \wedge  b_n(A_{n-1}, c')
\end{align*}
%
In contrast to binary rules, the head atom in unary rules contains a constant and only one instead of two variables. Such an expression can also be understood as a complex way to write down a unary predicate, which is the reason for naming these rules unary rules. Typical examples are head atoms such as $\textit{gender}(X,\textit{female})$ or $\textit{citizen}(X,\textit{spain})$.

We refer to rules of these types as path rules, because the body atoms form a \textit{path}. Note that our language bias also includes rule variations with flipped variables in the atoms: given a knowledge graph $\mathbb{G}$, a path of length $n$ is a sequence of $n$ triples $p_i(c_i, c_{i+1})$ with $p_i(c_i, c_{i+1}) \in \mathbb{G}$ or $p_i(c_{i+1}, c_i) \in \mathbb{G}$ for $0 \leq i \leq n$. The abstract rule patterns shown above are said to have a length of $n$ as their body can be instantiated to a path of length $n$. Instead of $A_i$ we will sometimes use $A$, $B$, $C$, and so on as names for the variables. Moreover, we will usually replace the variables that appear in the head by $X$ for the subject and $Y$ for the object.

\binary \ rules and \unarycon \ rules are also called closed connected rules. They can be learned by the mining system AMIE described in~\cite{amie2013,amieplus2015}. \unaryex \ rules are not closed because $A_n$ is a variable that appears only once.

Examples for binary rules are Rules~(\ref{rex1:r1}) and~(\ref{rex1:r2}) shown below. They describe the relation between $X$ and $Y$ via an alternative path between $X$ and $Y$. This path can contain a single relation or a chain of several relations. We allow recursive rules, i.e., the relation in the head can appear one or several times in the body as shown in Rule~(\ref{rex1:r2}). Rule~(\ref{rex1:r3}) is a \unarycon \ rule which states that a person is female, if she is married to a person that is male. A typical example for a \unaryex \ rule is Rule~(\ref{rex1:r4}), which says that an actor is someone who acts (in a film).
\begin{align}
\label{rex1:r1} \textit{hypernym}(X,Y)       & \leftarrow \textit{hyponym}(Y,X) \\
\label{rex1:r2} \textit{prod}(X,Y)       & \leftarrow \textit{prod}(X,A), \textit{sequel}(A,Y) \\
\label{rex1:r3} \textit{gen}(X,\textit{female})    & \leftarrow \textit{married}(X,A), \textit{gen}(A,\textit{male}) \\
\label{rex1:r4} \textit{prof}(X,\textit{actor}) & \leftarrow \textit{actedin}(X,A)
\end{align}

All considered rules are probabilistic which means they are annotated with confidence scores that represent the probability of predicting a correct fact with this rule. The fraction of body groundings that result in a correct head grounding (as measured on the training data) is called the \textit{confidence} of a rule. It is important to understand the relation between the three rule types. It is particularly interesting in the context of probabilistic rules. For that purpose, consider the following set of rules (fictitious confidence scores added in square brackets). 
\begin{align}
\label{rex2:r1} \textit{speaks}(X,Y)                  \leftarrow \textit{lives}(X,A), lang(Y,A)        & \ \ [0.8]  \\
\label{rex2:r2} \textit{speaks}(X,\textit{english})    \leftarrow \textit{lives}(X,\textit{A})    & \ \ [0.62] \\
\label{rex2:r3} \textit{speaks}(X,\textit{french})    \leftarrow \textit{lives}(X,\textit{france})     & \ \ [0.88] \\
\label{rex2:r4} \textit{speaks}(X,\textit{german})    \leftarrow \textit{lives}(X,\textit{germany})    & \ \ [0.95]
\end{align}
Let the relation $\textit{lives}(A,B)$ be used to say that a person $A$ lives in country $B$, and let $\textit{lang}(A,B)$ be used to say that a $A$ is (one of) the official languages of $B$. Thus, \binary\ rule~(\ref{rex2:r1}) states that $X$ speaks a certain language $Y$, if $X$ lives in a country $A$ where $Y$ is the official language. \unaryex\ Rule~(\ref{rex2:r2}) is a \textit{specialization} for predicting english speakers and the remaining \unarycon\ rules relate a specific language to a specific country. The interesting aspect of this rule set is the fact that Rule~(\ref{rex2:r2}) can be generated from Rule~(\ref{rex2:r1}) by removing the second atom in the body and by grounding $Y$ in the head. Likewise, Rules~(\ref{rex2:r3}) and (\ref{rex2:r4}) can be constructed by additionally grounding $A$. It seems that we do not need these specialized rule variants, if we already have a more \textit{general} rule. However, this is wrong for two reasons: (i) it might be the case that the given knowledge graph does not contain information about the official languages of France or Germany; and (ii) the confidences of the specific rules (\ref{rex2:r2})--(\ref{rex2:r4}) differ from the confidences of the more general rules. The confidence of a general rule is closely related to the (weighted) average over the specific confidences (e.g. by aggregating over all countries and languages). For that reason, it is necessary to generate both types of rules, even though they might carry partially redundant information.

\subsection{Sampling Rules}
\label{sub:learn-bottom}

We propose a bottom-up approach for learning rules from \textit{bottom rules}, i.e. grounded rules from sampled paths in the knowledge graph. It is divided into the following steps:
\begin{enumerate}
	\item Sample a path from a given knowledge graph.
	\item Construct a bottom rule from the sampled path.
	\item Build a generalization lattice rooted in the bottom rule.
	\item Store all useful rules that appear in the lattice.
\end{enumerate}
The above sketch of our approach reminds of the algorithm implemented in Aleph~\cite{aleph2000}. However, Aleph uses the bottom rule to define the boundaries of a top-down search. It begins with the most general rule and uses the atoms that appear in the bottom rule to create a specialization lattice. Similarly, AMIE also does a top-down search, which in contrast to Aleph is complete because it does not limit which atoms to use to specialize a rule. Our approach differs fundamentally from both algorithms because we create a generalization lattice beginning from the bottom rule. We argue in the following that all relevant rules within the generalization lattice instantiate one of the rule types defined in the previous section. Based on this insight, we can directly instantiate these rule types without the need to create the complete lattice.

\begin{figure}[ht!]
	\centering
	\includegraphics[width=0.85\columnwidth]{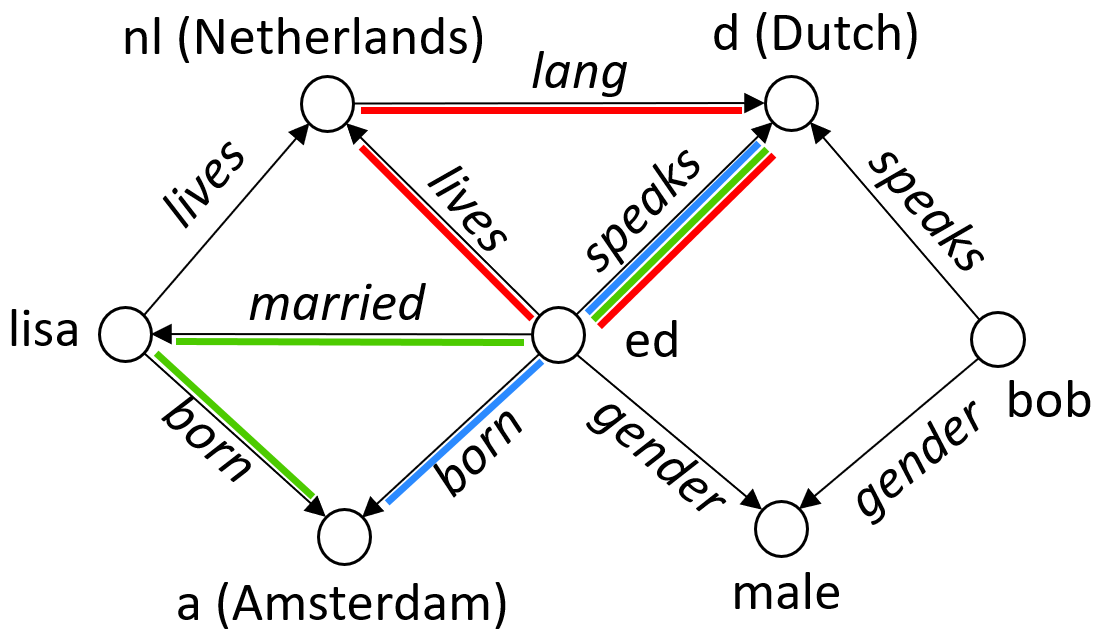}
	\caption{A knowledge graph $\mathbb{G}$ used for sampling paths. We marked the path that corresponds to Rule~\ref{rule:bottom-ex1} blue, Rule~\ref{rule:bottom-ex2} green, and Rule~\ref{rule:bottom-ex3} red.}
	\label{fig:kg}		
\end{figure}
To find rules for a fixed relation, AnyBURL samples multiple triples of that relation from the training set, and each time creates rules from it. Figure~\ref{fig:kg} shows a small subset of a knowledge graph $\mathbb{G}$. We use it to demonstrate how rules for the relation \textit{speaks} would be learned from it. We construct bottom rules of length $n$, beginning from $speaks(ed,d)$ (Ed speaks Dutch), which will be the head of the rules. To do this, we randomly walk $n$ steps in the graph, starting either from $ed$ or $d$. Together with the head triple, the result is a path of length $n + 1$. We have marked three paths in Figure~\ref{fig:kg} that could be found for $n=2$ or $n=1$, respectively. The green and blue paths are acyclic, while the red path, including $speaks(ed,d)$, is cyclic. We convert these paths into the bottom rules (\ref{rule:bottom-ex1}), (\ref{rule:bottom-ex2}), and (\ref{rule:bottom-ex3}).
\begin{align}
\label{rule:bottom-ex1} speaks(ed,d) \leftarrow & \ born(ed, a) \\
\label{rule:bottom-ex2} speaks(ed,d) \leftarrow & \ mar(ed, lisa), born(lisa, a) \\
\label{rule:bottom-ex3} speaks(ed,d) \leftarrow & \ lives(ed, nl), lang(nl, d)
\end{align}
We argue  that any generalization of a path of length $n+1$ will be a \binary, \unarycon \ or \unaryex \ rule of length $n$ or a shorter rule, which can be constructed from a shorter path, or a rule that is not useful for making a prediction. We elaborate this point by analysing the generalization lattice rooted in Rule~(\ref{rule:bottom-ex2}), depicted in Figure~\ref{fig:lattice-unary}.

Each edge in the lattice transition stems from one of the following two generalization operations. (i) Replace all occurrences of a constant by a fresh variable. (ii) Drop one of the atoms in the body. Note that we have only depicted those rules in the lattice that have at least one variable in the head. If this would not be the case, the rule would only predict a triple that is already stated in the knowledge graph, which is useless for completion. A rule that appears in the lattice falls into one of the following categories. We have associated the symbols $\dagger$, $\ast$, and $\diamond$ to each category and used them to mark the nodes in Figure~\ref{fig:lattice-unary}.
\begin{description}
	\item[Ambiguous prediction$^\dagger$] The rule has an unconnected variable in the head, which does not appear in the body of the rule. Such a rule makes a prediction that something exists, however, it does not make a concrete prediction which would be required to create a ranking of candidates.
	\item[Shorter bottom rule$^\ast$] The rule might be useful but it would also appear in the lattice of a bottom rule which originates from a shorter path. To avoid duplicate rules, we do not create it from the longer bottom rule. This point is detailed in Section~\ref{sec:search}.
	\item[Useless atom$^\diamond$] The body contains an atom without variables or an atom with a constant and an unbound variable. Such atoms will always be true in the knowledge graph from which they were sampled and therefore do not affect the truth value of the body. 
\end{description}
Note that a rule in the lattice marked with a $^\dagger$ or $^\ast$ does not need to be generalized any further, because any resulting rule will be marked again with the same symbol.
\pgfdeclarelayer{bg}    
\pgfsetlayers{bg,main}
 
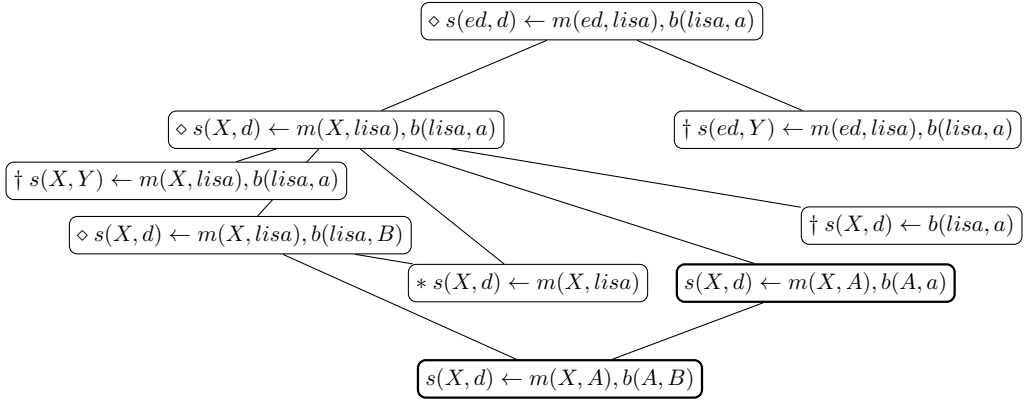
\begin{figure*}[t!]
    \centering
    \begin{tikzpicture}[scale=.85, transform shape]
    \tikzstyle{every node} = [rectangle, rounded corners=3pt, draw=black, fill=white]
    \coordinate (c1ffco) at (0.5,4.35);
    \node (a1) at (6,7.7) {$\diamond \ s(ed,d) \leftarrow m(ed,lisa), b(lisa,a)$};
    \node (b1) at (2,6) {$\diamond \ s(X,d) \leftarrow m(X,lisa), b(lisa,a)$};
    \node (b2) at (10,6) {$\dagger \ s(ed,Y) \leftarrow m(ed,lisa), b(lisa,a)$};
    \foreach \from/\to in {a1/b1, a1/b2} \draw [-] (\from) -- (\to);
    \node (c1) at (-0.5,5.2) {$\dagger \ s(X,Y) \leftarrow m(X,lisa), b(lisa,a)$};
    \node (c1ff) at (0.5,4.35) {$\diamond \ s(X,d) \leftarrow m(X,lisa), b(lisa,B)$};
    \node (c2) at (5,3.6) {$\ast \ s(X,d) \leftarrow m(X,lisa)$};
    \node[line width=0.3mm] (c3) at (9.5,3.6) {$s(X,d) \leftarrow m(X,A), b(A,a)$};
    \node (c4) at (11,4.5) {$\dagger \ s(X,d) \leftarrow b(lisa,a)$};
    \begin{pgfonlayer}{bg}
        \foreach \from/\to in {b1/c1, b1/c1ff, b1/c2, b1/c3, b1/c4} \draw [-] (\from) -- (\to);
        \node[line width=0.3mm] (d1) at (5.5,2.1) {$s(X,d) \leftarrow m(X,A), b(A,B)$};
        \foreach \from/\to in {c1ff/c2, c1ff/d1} \draw [-] (\from) -- (\to);
        \foreach \from/\to in {c3/d1} \draw [-] (\from) -- (\to);
    \end{pgfonlayer}
    \end{tikzpicture}
    \caption{Generalization lattice of the acyclic path $(s(ed,d), m(ed, lisa), born(lisa, a))$. For legibility we use the abbreviations $s=speaks$, $m=married$ and $b=born$.}
    \label{fig:lattice-unary}
\end{figure*}
%
%
When we apply this annotation scheme to the lattice (Figure~\ref{fig:lattice-unary}) that originates from the green acyclic path (in Figure~\ref{fig:kg}), only two rules remain unmarked. We have highlighted these rules with a bold rectangle. These two rules are of type~\unaryex \ and \unarycon. One can easily argue that this will always be the result when we generalize a bottom rule that originates from an acyclic path. Thus, we do not need to search over the generalization lattice but can directly create these two rules from a given acyclic path.

We can observe a similar pattern when we generalize a cyclic path. It results in three rules that we can leverage for a prediction; one \binary \ rule and two \unarycon \ rules, where the head constant (subject/object) appears again in the last body atom.


\subsection{Object Identity}
\label{sub:learn-identity}

Object Identity (OI) refers to an entailment framework that interprets every rule under the additional assumption that two different terms (variables or constants) that appear in a rule must refer to different entities. This means that each rule is extended by a pairwise complete set of inequality constraints. OI was first introduced in~\cite{semeraro1994avoiding} and later it is used to propose refinement operators for the original framework~\cite{esposito1996refinement}. In this work we do not focus on its theoretic properties but on its impact on correcting the confidence scores of the learned rules.

In the context of our approach, the most important property of OI is its capability to suppress redundant rules that negatively affect performance under the $\Theta$-subsumption~\cite{robinson1965machine} entailment regime. We illustrate the effect with the following two rules ($h$ and $b$ are two arbitrary but fixed relations).
\begin{align}
\label{rule:redrule1} h(X,Y) & \leftarrow h(X,Y) \\
\label{rule:redrule3} h(X,Y) & \leftarrow b(X,A), b(B,A), h(B,Y) 
\end{align}
Interpreting rules under OI can be done by adding additional constraints to the rules. For instance, the body of Rule~(\ref{rule:redrule3}) would need to be extended with the inequality constraints~(\ref{rule:constraints}).
\begin{align}
\label{rule:constraints} X \neq A, X \neq B, X \neq Y, A \neq B, A \neq Y, B \neq Y
\end{align}
Each rule constructed by AnyBURL is always interpreted under OI. Note that these inequality constraints are not shown whenever a rule is displayed or stored in a file.

Rule~(\ref{rule:redrule1}) is obviously a tautology that will never generate any new facts. This is only partially true for Rule~(\ref{rule:redrule3}). The groundings of its body can be divided into the groundings $\theta$ with $B = X$, and the groundings $\theta'$ with $B \neq X$. In contrast to a $\theta'$ grounding, a $\theta$ grounding does not predict new facts and is also more likely to result in a true body because both atoms of relation $b$ can be ground to the same fact. This means that, without OI, the confidence score of Rule~(\ref{rule:redrule3}) overestimates its quality as it will always be used to predict unknown facts. Adding the inequality constraints will suppress the $\theta$ groundings and result in a more realistic confidence score for the task. 

It is important to understand that it is not just variations of tautology rules that have this problem. For example, if there are strong rules such as $m(X,Y) \leftarrow spo(Y,X)$ (m = married, spo = spouse) in a knowledge graph, rules like the following are also affected.
\begin{align}
\label{rule:family} m(X,Y)\leftarrow son(X,A), son(B,A), spo(B,Y)
\end{align}
The confidence score of such a rule drastically (and rightfully) decreases under OI once we ignore groundings in which X and B are ground to the same son.



While OI helps us to avoid a blow-up of the rule base, a given rule is harder to evaluate under OI (see also §5.1.1 in~\cite{deraedt2008logical}). This holds both for the confidence computation as well as for the application of the rule in the context of predicting new knowledge. If we ignore the inequality constraints, all possible $(X,Y)$ groundings for Rule~(\ref{rule:redrule3}) can be computed with two join operations. As a result of the first join, we get the groundings for $(X,B)$ which can be used to compute the $(X,Y)$ groundings via a second join. However, the constraint $A \neq Y$ requires to know the variable bindings of $A$ that we used for the first join when doing the second join to ensure that the constraint is not violated. 
Keeping track of all variable bindings makes it more complex to compute body groundings under OI.

\section{Search Strategy}
\label{sec:search}


\subsection{Path Sampling}
\label{sub:search-pathsampling}

In the paths that we sample for building bottom rules, each triple on a path is called a step. The steps can be made in the direction of a stated triple or in reverse direction. A step in reversed direction causes flipped terms in the corresponding atom of the resulting rule. We call a path \textit{a straight path} if it does not visit the same entity twice, i.e., $c_i \neq c_j$ for each $i \neq j$.  An exception can be the equality of the first and last entity $c_0 = c_n$. In this case, we call the path \textit{a straight cyclic path}, which ends where it began.

A straight cyclic path results into a binary \binary\ rule and a special form of a \unarycon \ rule where the constant in the head and body of the rule is the same. A straight acyclic path results into a \unarycon \ rule (with different constants in head and body) and a \unaryex \ rule. Our method to sample a path is to choose a random entity as a starting point of a random walk. If the walk arrives at an entity that has been visited before (prior to the last step), the procedure can be restarted until a straight path has been found. This approach can yield cyclic and acyclic paths. It can be expected that the majority of sampled paths will be acyclic. Especially for longer paths it will not often be the case that $c_0 = c_n$. This means that a pure random walk strategy will generate only few binary rules. This can be a problem for the resulting rule sets. According to the results presented in~\cite{meilicke2018fine,meilicke2019anyburl} we know that a large fraction of correct predictions can be made with \binary \ rules. 

Thus, it makes sense to design a specific strategy to search for cyclic paths. We have slightly modified the random walk strategy by explicitly looking for a fact that connects $c_{n-1}$ and $c_0 = c_n$ in the last step. With an appropriate index it is possible to check the existence of a relation $p$ with $p(c_i, c_j)$ in constant time for any pair of constants. If we find such a triple, we use this as a final step in the constructed path. If we find several such triples, we pick randomly one of them. With this modification, we are able to find more cyclic paths in the same time span compared to the standard random walk. We are aware that there are more sophisticated methods for finding a path of length $n$, see for example~\cite{pallottino1984shortest}. 

\subsection{Saturation based Search}
\label{sub:search-saturation}


A detailed description of the search policy that was implemented in a previous version of AnyBURL can be found in~\cite{meilicke2019anyburl}. According to that policy, called saturation-based search, the learning process is conducted in a sequence of time spans of fixed length (e.g. one second).  Within a time span the algorithm learns as many rules as possible using paths sampled from a specific path profile. A path profile describes path length and whether the path is cyclic or acylic. When a time span is over, the rules found within this span are evaluated. Let $R$ refer to the rules that have been learned in the previous time spans, let $R_s$ refer to the rules found in the current time span, and let $R_s' = R_s \cap R$ refer to the rules found in the current time span that have also been found in one of the previous iterations.  If $|R_s'| / |R_s |$ is above a saturation boundary, which needs to be defined as a parameter, the path length of the profile is increased by one. Initially, the algorithm starts with paths of length 2 resulting in rules of length 1. The higher the path length, the more time spans are usually required to reach the saturation boundary. The difference between cyclic and acyclic paths is taken into account by flipping every time span between the cyclic and  acyclic profiles. The rule counts generated by cyclic and acyclic rules are independent. 

It is an advantage of the saturation-based approach that it does not require to mine all cyclic (or acyclic) rules of length $n$ before the algorithm looks at cyclic (or acyclic) rules of length $n+1$. Instead of that, the rule length is increased if a sufficient saturation has been reached. However, it is unclear how to set the required saturation degree (the default value is 0.99). If this value is set too high, the algorithm spends a lot of time sampling paths resulting into rules that have already been generated previously. If the value is set too low, important rules might be missed and cannot be found any more. Another disadvantage of the algorithm is that the ad hoc setting to spend exactly half of the time to search for cyclic paths and half for acyclic paths. 
To overcome these shortcomings we propose a reinforced approach presented in the following section.

\subsection{Reinforced Search}
\label{sub:search-reinforced}

\subsubsection{Reward}
In the following we consider the path  sampling problem as a special kind of multi-armed bandit problem~\cite{katehakis1987multi}. In each time span we have to decide how much effort to spend on which path profile. A path profile in our scenario corresponds to an arm of a bandit in the classical reinforcement learning setting. Each arm (or slot machine) in the bandit problem gives a reward when pulling that arm. What corresponds in our scenario to the reward of pulling an arm, i.e., the reward of creating rules from the paths that belong to a certain profile?

In the following we develop three different reward strategies. They are based on the notion of measuring the reward paid out by a profile in terms of the explanatory quality of the rules that were created by that profile. The explanatory quality of a rule set can be measured in terms of the number of triples of a given knowledge graph that can be reconstructed with the help of the rules from the set. Thus, summing up the support of the rules seems to be a well suited metric. We refer to this as reward strategy $R_s$. 
\begin{equation}
R_s(\mathbb{S}) = \sum_{r \in \: \mathbb{S}} \mathit{support}(r) 
\end{equation}
where $\mathbb{S}$ is a set of rules and $support(r)$ is the support of a rule $r$. Given a rule $r = \hat{r} \leftarrow \check{r}$,  we denote by 
$r\theta_ X$ the (partially) grounded rule where all occurrences of $X$ are
replaced by some constant. Consequently, support and confidence of $r$ can be defined as follows:
$$\mathit{support}(r) = \left| \{ \theta_{XY}  \mid \exists \theta_{Z} \ \check{r}\theta_{XYZ} \wedge \hat{r}\theta_{XY}\} \right|$$

$$\mathit{conf}(r) = \frac{\left| \{ \theta_{XY}  \mid \exists \theta_{Z} \ \check{r}\theta_{XYZ} \wedge \hat{r}\theta_{XY}\} \right|}{\left| \{ \theta_{XY}  \mid \exists \theta_{Z} \ \check{r}\theta_{XYZ} \} \right|}$$
where $\theta_{XY}$ refers to a grounding for variables $X$ and $Y$, which appear in the head of $r$. $\theta_{Z}$ is a grounding for the variables that appear in the body of $r$ that are different from $X$ and $Y$. $\theta_{XYZ}$ refers to the union of $\theta_{XY}$ and $\theta_{Z}$.

We are especially interested in rules that make many correct predictions with high confidence. Since, predictions with high confidence are more likely to appear as a top ranked candidate. For that reason we propose a second reward strategy $R_{s \times c}$ that multiplies the number of correct predictions by their confidence:
\begin{equation}
R_{s \times c}(\mathbb{S}) = \sum_{r \in \: \mathbb{S}} \mathit{support}(r) \times\mathit{conf}(r)
\end{equation}
where $\mathbb{S}$ is a set of rules, $support(r)$ is the support and $conf(r)$ is the approximate confidence of a rule $r$.



We define a third reward strategy that takes rule length into account as follows 
\begin{equation}
R_{s \times c / 2^l}(\mathbb{S}) = \sum_{r \in \: \mathbb{S}} \mathit{support}(r) \times\mathit{conf}(r) / 2^{l(r)} 
\end{equation}
where $l(r)$ denotes the length of a rule $r$. This reward strategy is  a variant of $R_{s \times c}$ that favours shorter over longer rules. It enforces a constraint that assigns at the beginning of the search more computational effort to short rules. Thus, the search constraint has some similarities with a softened saturation-based search as long as we are only concerned with rule length.

All metrics are based on the capability of a rule set to reconstruct parts of a given knowledge graph in terms of the training set. An alternative approach would have been to compute the same or similar scores with respect to the prediction of the validation set. If we focus on the training set, we can directly reuse the scores that we already computed. Additional computational effort is not required.


\subsubsection{Policy}
All three reward strategies can be combined with each of the following two policies. The first policy is a well known policy referred to as $\epsilon$-greedy policy~\cite{sutton2018reinforcement}. The parameter $\epsilon$ is usually set to relatively small positive value, for example $\epsilon = 0.1$. Every time a decision needs to be made, that decision is a random decision with a probability $<$~$\epsilon$ and a greedy decision with a probability $\geq$~$\epsilon$. When we talk about decisions, we mean the allocation of CPU cores to path profiles. In the $\epsilon$-greedy policy, a small number of decisions is randomized to reserve a small fraction of the available resources for exploration compared to an approach that would focus completely on exploitation.

In our context, a greedy decision assigns all cores, that have not been assigned randomly, to the path profile that generated the rule set that yielded the highest reward the last time it has been selected. Formally, for $\epsilon$-greedy policy, a path profile $\mathit{pf}^*$, with $1-\epsilon$ probability, is chosen for time span $t_k$ according to the following equations
\begin{align*}
\mathit{pf}^* =&~ \operatorname*{argmax}_{\mathit{pf}\in \mathbb{F}} Q(\mathit{pf, last(pf, t_k)}), \\
Q(\mathit{pf,t_i}) =&~ \frac{1}{N_{t_i}(\mathit{pf})} \mathbf{R}(\mathit{S(pf, t_i)} \setminus \bigcup_{j=1}^{i-1} \mathit{S(pf, t_j)})
\end{align*}
where $\mathit{last(pf, t_k)}$ refers to the last time span $t_i$ prior to $t_k$ (i.e., with $i<k$) where path profile $\mathit{pf}$ has been used, $Q(\mathit{pf, t_i})$ is the value of the path profile $\mathit{pf}$ for time span $t_i$, $N_{t_i}(\mathit{pf})$ quantifies the computational resources that have been allocated to $\mathit{pf}$ during $t_i$,  $\mathbf{R}$ denotes a reward strategy $R_s$, $R_{s \times c}$ or $R_{s \times c / 2^l}$, and $\mathit{S(pf, t_i)}$ refers to the set of rules that have been mined by the use of path profile $\mathit{pf}$ during $t_i$. The expression $\mathit{S(pf, t_i)} \setminus \bigcup_{j=1}^{i-1} \mathit{S(pf, t_j)}$ refers to the set of new rules that have been mined in $t_i$ but not in one of the previous time spans. We quantify $N_{t_i}(\mathit{pf})$ in terms of the number of cores that are assigned to $\mathit{pf}$ during $t_i$. This means that the reward is normalized with the number of allocated cores. 

Note that our scenario differs from the classical multi-armed bandit setting in the sense that the expected reward of a certain profile will decrease any time we use this profile for generating rules.
The more often we use that profile, the more probably it is to draw a path that results into a previously learned rule, which was created from the same or from a different path. For that reason we do not base our decision on the average over all previous time spans, but look at the last time span that this profile has been used. The reward of a profile is shrinking continuously, with random ups and downs that are caused by drawing only a limited number of samples. This results into flips between different profiles that are not caused by knowing more (exploration) but by the impact of exhausting profiles over time.

The $\epsilon$-greedy policy might not be a good choice if one profile $\mathit{pf}$ creates higher rewards than another profile $\mathit{pf}'$, however, $\mathit{pf}'$ would also generate relatively good rules. 
Suppose further that both profiles are relatively stable, i.e., their reward decreases only slightly when they are used for generating rules. In such a setting, we might prefer to draw rules not only from $\mathit{pf}$ but also from $\mathit{pf}'$. For that reason we propose a second policy where we distribute the available computational resources to all profiles proportional to the reward that has been observed the last time they have been used. We refer to this policy as weighted policy.
For each CPU core with a probability $<$~$\epsilon$, we take  a random decision; and with a probability $\geq$~$\epsilon$ we proceed as follows. For each profile $\mathit{pf} \in \mathbb{F}$ we compute the probability of resource allocation $P_k(\mathit{pf})$ at time span $t_k$, given by the following formula: 
\begin{align*}
P_{k}(\mathit{pf}) =&~ \frac{Q(\mathit{pf, last(pf, t_k)}) }{\sum_{\mathit{pf}' \in \mathbb{F}}^{} Q(\mathit{pf, last(pf', t_k)})}
\end{align*}
where $Q$ and $last$ are introduced above. For each core that is not yet assigned to a profile due to the random assignment in the $<$$\epsilon$ case, we throw a dice and assign one of the path profile $\mathit{pf} \in \mathbb{F}$ with probability $P_{k}(\mathit{pf})$.


To better understand the impact of combining different reward strategies and policies, AnyBURL can be run with a completely random policy where each profile has always the same probability. This can be achieved by setting $\epsilon = 1$. This setting is not necessarily bad. If there are $K$ different profiles, in the worst scenario one of these profiles would generate many useful rules and none of the other profiles would generate such rules. An algorithm that makes perfect decisions would arrive $K$ times faster at the same result as the random policy. However, at the same time we can assume -- and the results published in~\cite{meilicke2019anyburl} support this assumption -- that the most beneficial rules are often mined first. This means that running the random policy and the weighted policy for the same time span will not yield results that are $K$ times worse. The random policy might even outperform the previous, saturation-based implementation of AnyBURL. This will be the case if the saturation threshold is chosen too low or too high. 


\section{Experiments}
\label{sec:experiments}

\subsection{Datasets and Settings}
\label{sub:exp-datasets}

We use in our experiments the datasets FB15(k), its modified variant FB15-237, WN18, and its modified variant WN18RR. The FB (WN) datasets are based on a subset of FreeBase (WordNet). FB15 and WN18 have been first used in~\cite{transe2013}. They have been criticised in several papers~\cite{toutanova2015observed,conve2018}, where the authors argued that due to redundancies a large fraction of testcases can be solved by exploiting rather simple rules. FB15-237~\cite{toutanova2015observed} and WN18RR~\cite{conve2018} have been proposed as modified variants with suppressed redundancies. 
The dataset YAGO03-10 (in short YAGO) is described in~\cite{mahdisoltani2015yago3} and has first been used in the context of knowledge completion in~\cite{conve2018}. 
It is two times larger in number of triples compared to FB15. An overview is given in Table~\ref{tab:datasets}.

\begin{table}
	\centering
\resizebox{\columnwidth}{!}{%
		\begin{tabular}{l|r|r|r|r|r}
\toprule
		         &  WN18   & WN18RR  & FB15     & FB15-237  & YAGO03-10  \\
\midrule
Entities     &  40943  & 40559   &  14951   &  14505    &  123143   \\
Relations    &     18  &    11   &   1345   &    237    &      37   \\
Triples      & 141442  & 86835   & 483142   & 272115    & 1079040  \\
Testset      &   5000  &  3134   &  59071   &  20466    &    5000  \\
\bottomrule
		\end{tabular}}
	\caption{Dataset characteristics. The numbers reported in the first three lines refer to the training set. Each triple in the test set (fourth row) can be divided into two test cases.}
  \label{tab:datasets}
\end{table}

The most commonly used evaluation metrics are the filtered hits@1 and hits@10 scores introduced in~\cite{transe2013}. The hits@k scores measure how often (fraction of test cases)  the correct answer, which is defined by the triple that the test case originates from, is among the top-k ranked entities. 
In the following we always refer to the filtered scores without explicitly stating it. 
Another important value is the filtered MRR (mean rank reciprocal). As our approach is not designed to compute complete rankings but top-k rankings only, we compute a lower bound by assuming that any candidate which would be ranked at a position $>$k is not a correct prediction.


As described in \cite{meilicke2019anyburl}, we use max aggregation to generate predictions from the rule set. We learn rules up to length 3 from cyclic paths (length 5 for WN and WN18RR) and restrict the length of rules learned from acyclic paths to 1. Confidences of rules are approximated by sampling and evaluating groundings on the training set followed by a laplace smoothing with parameter $p_c=5$. We keep all rules with a confidence higher than 0.0001 that reconstructed at least two triples in the training set. If not stated otherwise, we use the weighted reinforced policy together with reward strategy $R_{s \times c}$. We are running AnyBURL on a CPU sever with 24 Intel(R) Xeon(R) CPU E5-2630 v2 @ 2.60GHz cores. We use 22 threads 
and reserve 50 GB RAM for our experiments. 

\subsection{Object Identity}
\label{sub:exp-oi}

We first run AnyBURL with deactivated OI constraints for a fixed amount of time (1000 seconds) on the WN18 dataset. For that purpose we deactivate rules with constants and learn only binary rules of length 1 to 5. We run our experiments in two settings. In a strict setting, we set the minimum support to 100 and the minimum confidence to 0.5. In a relaxed setting, we use lower thresholds, i.e., we set the minimum support to 10 and the confidence threshold to 0.1. In a post processing step, we activate the OI constraints and recompute confidences for the previously computed rule sets. Then we count the fraction of rules that remain above these thresholds. We evaluate both rules sets in both settings on the test set to measure the quality of the resulting predictions.

To avoid inaccuracies caused by approximated confidences, we filter only those rules for which the scores are significantly lower than the threshold. Otherwise, we would not know if a rule that was only slightly above the threshold falls below the threshold due to a sampling inaccuracy or due to an OI constraint. We assume that a confidence or support score is significantly lower if the value is lower than half of the originally chosen threshold. 

\begin{table}
	\centering	
\resizebox{\columnwidth}{!}{	
		\begin{tabular}{c|c|c|c|c|c}
\toprule		
\textbf{Thresholds}                                   & \textbf{OI}   & \textbf{Rules} & \textbf{Reduced} & \textbf{Hits@1} & \textbf{Hits@10} \\
\midrule

\multirow{2}{*}{s$\geq$100, c$\geq$0.5} & off  & 2004  & \multirow{2}{*}{10.7\%}  & 0.739 & 0.88 \\
                                        & on   &  215  &                          & 0.938 & 0.942 \\
\midrule
\multirow{2}{*}{s$\geq$10, c$\geq$0.1}  & off &  12832 & \multirow{2}{*}{58.3\%}  & 0.765 & 0.88 \\
                                        & on   & 7475  &                          & 0.944 & 0.957 \\
\bottomrule													
		\end{tabular}}
	\caption{Comparing the impact of OI constraints on two different rule sets for the WN18 dataset, $s$ and $c$  denote support and confidence scores respectively.}
	\label{tab:oiexp}
\end{table}

The results of our experiments are shown in Table~\ref{tab:oiexp}. We can see that activating OI constraints has a strong impact on rules that have a high confidence without these constraints. This is highlighted by the results for the strict setting, shown in the first two rows. The number of rules drops from 2004 to 215, which means that only 1/10 of the rules remain above the thresholds. Furthermore, the predictive results support the theoretic considerations from Section~\ref{sub:learn-identity}. The fact that hits@1 increases from 0.739 to 0.938, is a strong indicator that many of the rules, which have been filtered, had a confidence score that was too high without OI constraints. It is also important to note that OI constraints are not only useful to obtain a high precision, which is reflected in the hits@1 score, but also we  observe a significant improvement for hits@10.

The impact on filtering is less strict for the second setting. Around half of the rules are filtered out. However, the chosen thresholds are relatively low. Nevertheless, the impact on the predictive quality is similar to the first setting. This is caused by the modified confidence scores. The results are not surprising, because the rule set generated in the first setting is the subset of the second rule set that includes the most influential rules. However, if we compare both settings under OI, the second setting achieves better hits@1 and hits@10 scores. This means, that rules with a confidence lower than 0.5 are now able to contribute to the generated ranking.

As argued in several studies~\cite{toutanova2015observed,conve2018,rulen2018}, WN18 allows to learn many simple rules that have a high predictive power. These rules can then become redundant building blocks in longer rules. As soon as we interpret these rules under OI, their scores are corrected, resulting in better predictions and, if we apply a threshold, into smaller rule sets. While the impact might be less strong on other datasets, the underlying pattern will always have an impact as long as there are rules of different length and some of the shorter rules have a relatively high score. 

\subsection{Reinforcement Learning}
\label{sub:exp-reinforced}

\begin{figure}
\begin{center}
\resizebox{\columnwidth}{!}{
\begin{tikzpicture}[scale=1.5,cap=round]

\draw[-] (-0.1,0) -- (5.2,0);
\draw[-] (0,-0.2) -- (0,2.2);

\draw (0.25,-0.2) node {$50$s};
\draw (1,-0.2) node {$200$s};
\draw (2,-0.2) node {$400$s};
\draw (3,-0.2) node {$600$s};
\draw (4,-0.2) node {$800$s};
\draw (5,-0.2) node {$1000$s};

\draw (0,2.4) node {hits@10};
\draw (-0.31,0) node {0.6};
\draw (-0.41,0.5) node {0.625};
\draw (-0.31,1) node {0.65};
\draw (-0.41,1.5) node {0.675};
\draw (-0.31,2.0) node {0.7};
\draw[gray,very thin] (-0.07, 0.5) -- (0.07, 0.5);
\draw[gray,very thin] (-0.07, 1.0) -- (0.07, 1.0);
\draw[gray,very thin] (-0.07, 1.5) -- (0.07, 1.5);
\draw[gray,very thin] (-0.07, 2.0) -- (0.07, 2.0);


\draw[gray,very thin] (-0.07,0.5) -- (5.2,0.5);
\draw[gray,very thin] (-0.07,1) -- (5.2,1);
\draw[gray,very thin] (-0.07,1.5) -- (5.2,1.5);
\draw[gray,very thin] (-0.07,2) -- (5.2,2);

\draw[gray,very thin] (0.25,-0.07) -- (0.25,0.07);
\draw[gray,very thin] (1.0,-0.07) -- (1.0,2.2);
\draw[gray,very thin] (2.0,-0.07) -- (2.0,2.2);
\draw[gray,very thin] (3.0,-0.07) -- (3.0,2.2);
\draw[gray,very thin] (4.0,-0.07) -- (4.0,2.2);
\draw[gray,very thin] (5.0,-0.07) -- (5.0,2.2);

\draw[black,thick] plot file {data/yago-c3a1-sat09};
\draw[purple,thick] plot file {data/yago-c3a1-sat099};
\draw[violet,thick] plot file {data/yago-c3a1-sat0999};
\draw[red,dashed,thick] plot file {data/yago-c3a1-rand};
\draw[orange, thick, dotted] plot file {data/yago-c3a1-p2s3};

\draw[black, thick] plot coordinates {(5.4,2.2) (6.0,2.2)};
\draw[black] (5.7,2.25) node[anchor=north] {\footnotesize{sat$_{0.9}$}};

\draw[purple, thick] plot coordinates {(5.4,1.7) (6.0,1.7)};
\draw[black] (5.7,1.75) node[anchor=north] {\footnotesize{sat$_{0.99}$}};

\draw[violet, thick] plot coordinates {(5.4,1.2) (6.0,1.2)};
\draw[black] (5.7,1.25) node[anchor=north] {\footnotesize{sat$_{0.999}$}};

\draw[red, thick, dashed] plot coordinates {(5.4,0.6) (6.0,0.6)};
\draw[black] (5.7,0.65) node[anchor=north] {\footnotesize{rand}};

\draw[orange, thick, dotted] plot coordinates {(5.4,0.0) (6.0,0.0)};
\draw[black] (5.7,0.05) node[anchor=north] {\footnotesize{$R_{s \times c}$}};

\end{tikzpicture}}
\end{center}
\caption{Results for the YAGO dataset comparing saturation based approach with three different saturation boundaries (0.9, 0.99, 0.999) against random and weighted policy using $R_{s \times c}$.}
\label{fig:riff-all-yago}
\end{figure}
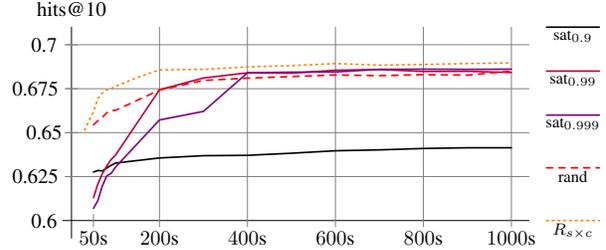

We used the largest dataset YAGO03-10 to compare the (i) saturation based approach with different saturation boundaries (0.9, 0.99, and 0.999) against the (ii)  random policy and the (iii) weighted reinforcement policy together with $R_{s \times c}$ in a first experiment. We learned rules in each of these settings for 1000 seconds, taking frequent snapshots of the learned rules. For each of these snapshots, we computed the predictions against the test set. The resulting hits@10 scores are depicted in Figure~\ref{fig:riff-all-yago}. We observe a very good anytime behaviour for most settings. The weighted policy causes the fastest increase: after 200 seconds we have learned a rule set with a hits@10 score of 68.6\%. This score is only slightly improved by 0.4\% when increasing the available time to 1000 seconds. The second best approach is the random policy, if we consider a quick improvement at the beginning as important. After a short time (50 to 200 seconds), it achieves better results than the  saturation based approach with a boundary set to 0.99. There is also a time period in which a saturation-based approach performs slightly better. 

We observed for all three settings of the saturation-based approach, that the saturation for \unaryex\ and \unarycon\ rules of length one created from acyclic paths does not reach the boundary within 1000 seconds. This is different for the rules generated from cyclic paths. The saturation boundary of 0.9 has been passed after 4 seconds for rules of length one and again after 9 seconds for rules of length two. The corresponding times were 16 and 198 seconds for 0.99, and 307 and 397 seconds for 0.999 respectively.

A saturation boundary of 0.9 is too low. The early jump to the longer paths causes that some beneficial rules cannot be found. This seems not to be the case for the boundary 0.99. However, the results for 0.99 are slightly worse than the results for 0.999 after mining rules for 1000 seconds. While a high boundary of 0.999 is beneficial on the long run, it prevents that good scores are achieved early. This is caused by the importance of some \binary\ rules of length three, which are not generated before 397 seconds have passed.

\begin{table}
	\centering
\resizebox{\columnwidth}{!}{
\begin{tabular}{cc|c|rrr|rrr}
\toprule
&  \textbf{Time(s)} &   \textbf{Random} &  \multicolumn{3}{c|}{\textbf{Greedy}} &  \multicolumn{3}{c}{\textbf{Weighted}}   \\ 
&       &          & $R_{s}$ \ \ & $R_{s\times c}$  & $R_{s \times c / 2^l}$ & $R_{s}$ \ \ &  $R_{s \times c}$  & $R_{s \times c / 2^l}$  \\ 
\midrule
\multirow{4}{*}{\rotatebox{90}{\footnotesize{\textbf{FB15-237}}}}
 & 50   & 47   & +0.6  & +0.5  & +0.8  & +0.5  & +0.8  & +0.6 \\
 & 100  & 48.5 & +0.8  & +0.3  & +0.8  & +0.6  & +0.8  & +0.7 \\
 & 500  & 51   & +0.6  & +0.4  & +0.7  & +0.6  & +0.5  & +0.6 \\
 & 1000 & 51.5 & +0.4  & +0.3  & +0.5  & +0.5  & +0.3  & +0.4 \\
\midrule
\multirow{4}{*}{\rotatebox{90}{\footnotesize{\textbf{FB15}}}}
 & 50   & 86.9 & -1.7 & -0.6 & +0.4 & -2.3 & -1.3 & -1.5 \\
 & 100  & 88.4 & -0.5 & -0.4 &   0  & -0.3 & -0.2 &    0 \\
 & 500  & 89.1 & -0.1 & -0.3 & +0.1 & +0.1 &    0 & -0.1 \\
 & 1000 & 89.2 & -0.1 & -0.3 & 0    &    0 &    0 & -0.1 \\

\midrule
\multirow{4}{*}{\rotatebox{90}{\footnotesize{\textbf{YAGO3-10}}}}
 & 50   & 65   & +0.8  & +0.8  & +0.3  & +0.9 & +0.4 & +1.0   \\
 & 100  & 65.7 & +1.4  & +1.4  & +1.1  & +1.2 & +1.2 & +1.4 \\
 & 500  & 67.8 & +0.7  & +0.6  & +0.6  & +0.6 & +0.6 & +0.7 \\
 & 1000 & 68.3 & +0.5  & +0.5  & +0.3  & +0.4 & +0.4 & +0.4 \\
\bottomrule
		\end{tabular}}
			\caption{Comparing policies and reward strategies against the random policy in terms of hits@10.}
	\label{tab:results-reinforced}
\end{table}

In the following, we compare the random baseline against all possible combinations of policies and reward strategies. Results are shown in Table~\ref{tab:results-reinforced}. We evaluated each setting three times (six times for 50s and 100s) and report the  resulting averages. In particular, we compare each reinforcement setting against the random policy on the three largest datasets showing the difference in terms of reinforced versus random approach. We observe improvements compared to the random policy for FB15-237 and YAGO for each combination of policy and reward strategy. However, there is not a single combination that performs clearly better than the other ones. This is a bit surprising, as we would have expected a positive impact of taking confidence into account.

To better understand the meaning of the numbers in Table~\ref{tab:results-reinforced}, we take a look at the hits@10 gain of +0.6\%  in the \textit{Weighted}/$R_{s \times c}$ column of the YAGO03-10 \textit{500 seconds} row. A plus of 0.6\% looks like a minor improvement at first sight. However, changing from  random to the reinforcement policy achieves $67.8\% + 0.6\% = 68.4\%$ hits@10 which is higher than the 1000 seconds score of the random policy (68.3\%). Thus, the same (or slightly better) results are achieved in half of the time.

The results for FB15 after 50 and 100 seconds are an exception from this trend. All policies guide the search into the wrong direction at the beginning and the reward strategies, that do not take into account rule length, perform also worse after 100 seconds. After 500 seconds the random policy and the other policies achieve similar results. A possible explanation for this behaviour can be the fact that FB15 has many redundancies and a high number of relations. This implies that regularities that require longer rules can be expressed in many different ways by replacing one atom by an (nearly) equivalent atom. For that reason, cyclic path profiles of length two and three receive a reward by $R_{s}$ and $R_{s\times c}$ that is too high compared to short rules. This is also the reason why the greedy policy together with reward strategy  $R_{s \times c / 2^l}$ performs best on FB15. It favours short rules over longer rules and mitigates the described effect without negative impact on the results measured for the other datasets.

\subsection{State of the Art}
\label{sub:exp-stateoftheart}

In the first block of Table~\ref{tab:main} we compare the results of AnyBURL against 16 different models presented in~\cite{rossi2020knowledge}. The second block (marked with $\star$) lists the results from~\cite{ruffinelli2020dog}. Here the authors report about the performance of the classic models RESCAL, TransE, DistMult, ComplEx and ConvE, arguing that these models perform better than usually reported and quite comparable to each other if the training strategies and other relevant hyperparameters are correctly tuned. While~\cite{rossi2020knowledge} reports numbers for all datasets that we used, \cite{ruffinelli2020dog}~report only results related to WN18RR and FB15-237. 
 In the AnyBURL block, the $\dagger$10000s row refers to the 10000 seconds run of the previous AnyBURL version reported in~\cite{meilicke2019anyburl}, while the rows below refer to the new version. 

\begin{table*}
	\centering
\resizebox{\textwidth}{!}{\begin{tabular}{cl|rrr|rrr|rrr|rrr|rrr}
\toprule
   &        & \multicolumn{3}{c|}{\textbf{FB15}} & \multicolumn{3}{c|}{\textbf{WN18}} & \multicolumn{3}{c|}{\textbf{FB237}} & \multicolumn{3}{c|}{\textbf{WN18RR}} & \multicolumn{3}{c}{\textbf{YAGO03-10}} \\
 & Approach & h@1   & h@10  & MRR   &	h@1  & h@10  & MRR   & h@1   & h@10  & MRR   & h@1   & h@10  & MRR & h@1   & h@10  & MRR   \\
\midrule
\multirow{9}{*}{\rotatebox{90}{\cite{rossi2020knowledge}}}
 & ConvE  & 59.46 & 84.94 & 0.688 & 93.89 & 95.68 & 0.945 & 21.90  & 47.62 & 0.305 & 38.99 & 50.75 & 0.427 & 39.93 & 65.75 & 0.488 \\
 & ConvKB & 11.44 & 40.83 & 0.211 & 52.89 & 94.89 & 0.709 & 13.98 & 41.46 & 0.23  & 5.63  & 52.5  & 0.249 & 32.16 & 60.47 & 0.42  \\
 & ConvR & 70.57 & 88.55 & 0.773 & 94.56 & 95.85 & 0.95  & 25.56 & 52.63 & 0.346 & 43.73 & 52.68 & 0.467 & 44.62 & 67.33 & 0.527 \\
 & CapsE  & 1.93  & 21.78 & 0.087 & 84.55 & 95.08 & 0.89  & 7.34  & 35.60  & 0.16  & 33.69 & 55.98 & 0.415 &       &       &       \\
 & RSN    & 72.34 & 87.01 & 0.777 & 91.23 & 95.10  & 0.928 & 19.84 & 44.44 & 0.28  & 34.59 & 48.34 & 0.395 & 42.65 & 66.43 & 0.511 \\
\cline{2-17}
 & TransE  & 49.36 & 84.73 & 0.628 & 40.56 & 94.87 & 0.646 & 21.72 & 49.65 & 0.31  & 2.79  & 49.52  & 0.206 & 40.57 & 67.39 & 0.501 \\
 & STransE & 39.77 & 79.60  & 0.543 & 43.12 & 93.45 & 0.656 & 22.48 & 49.56 & 0.315 & 10.13 & 42.21 & 0.226 &  3.28 &  7.35 & 0.049 \\
 & CrossE  & 60.08 & 86.23 & 0.702 & 73.28 & 95.03 & 0.834 & 21.21 & 47.05 & 0.298 & 38.07 & 44.99 & 0.405 & 33.09 & 65.45 & 0.446 \\
 & TorusE  & 68.85 & 83.98 & 0.746 & 94.33 & 95.44 & 0.947 & 19.62 & 44.71 & 0.281 & 42.68 & 53.35 & 0.463 & 27.43 & 47.44 & 0.342 \\
 & RotatE  & 73.93 & 88.10  & 0.791 & 94.30  & 96.02 & 0.949 & 23.83 & 53.06 & 0.336 & 42.60  & 57.35 & 0.475 & 40.52 & 67.07 & 0.498 \\
\cline{2-17}
 & DistMult & 73.61 & 86.32 & 0.784 & 72.6  & 94.61 & 0.824 & 22.44 & 49.01 & 0.313 & 39.68 & 50.22  & 0.433 & 41.26 & 66.12 & 0.501 \\
 & ComplEx  & 81.56 & 90.53 & 0.848 & 94.53 & 95.5  & 0.949 & 25.72 & 52.97 & 0.349 & 42.55 & 52.12 & 0.458 & 50.48 & 70.35 & 0.576 \\
 & ANALOGY  & 65.59 & 83.74 & 0.726 & 92.61 & 94.42 & 0.934 & 12.59 & 35.38 & 0.202 & 35.82 & 38.00    & 0.366 & 19.21 & 45.65 & 0.283 \\
 & SimplE   & 66.13 & 83.63 & 0.726 & 93.25 & 94.58 & 0.938 & 10.03 & 34.35 & 0.179 & 38.27 & 42.65 & 0.398 & 35.76 & 63.16 & 0.453 \\
 & HolE     & 75.85 & 86.78 & 0.8   & 93.11 & 94.94 & 0.938 & 21.37 & 47.64 & 0.303 & 40.28  & 48.79 & 0.432 & 41.84 & 65.19 & 0.502 \\
 & TuckER   & 72.89 & 88.88 & 0.788 & 94.64 & 95.8  & 0.951 & 25.90  & 53.61 & 0.352 & 42.95 & 51.40  & 0.459 & 46.56 & 68.09 & 0.544 \\
\midrule
\multirow{5}{*}{\rotatebox{90}{($\star$)}}
 & RESCAL   &  &  &  &  &  &  &                              26.3  & 54.1 & 0.357  & 43.9  & 52.1  & 0.468 & & & \\
 & TransE   &  &  &  &  &  &  &                              22.1  & 49.7 & 0.313  &  5.3  & 52.6  & 0.227 & & & \\
 & DistMult &  &  &  &  &  &  &                              25.0  & 53.1 & 0.343  & 41.3  & 53.1  & 0.452 & & & \\
 & ComplEx  &  &  &  &  &  &  &                              25.3  & 53.4 & 0.348  & 43.8  & 54.3  & 0.477 & & & \\
 & ConvE    &  &  &  &  &  &  &                              24.8  & 52.1 & 0.339  & 41.1  & 50.8  & 0.447 & & & \\
\midrule
\multirow{5}{*}{\rotatebox{90}{AnyBURL}}  
& $\dagger$10000s  & 80.4   & 89.0 & $\geq$0.83 & 94.6 & 95.9 & $\geq$95 & 23.3 & 48.6 & $\geq$0.31 & 44.1  & 55.2  & $\geq$0.47  & 47.7 & 67.3 & $\geq$0.54 \\ 
\cline{2-17}
& 100s             & 80.14 & 87.39 & $\geq$0.823 & 94.75 & 96.13 & $\geq$0.952 & 25.40  &  48.54 & $\geq$0.322 & 45.49 & 57.42 & $\geq$0.49   & 48.46 & 66.63  & $\geq$0.542  \\ 
& 1000s            & 81.44 & 89.42 & $\geq$0.839 & 94.76 & 96.17 & $\geq$0.952 & 27.34 &  52.25 & $\geq$0.346 & 45.69 & 57.67 & $\geq$0.492  & 49.24 & 68.94  & $\geq$0.555  \\  
& 10000s           & 81.02 & 89.39 & $\geq$0.836 & 94.79 & 96.12 & $\geq$0.952 & 27.20  &  52.03 & $\geq$0.345 & 45.37 & 57.24 & $\geq$0.488  & 49.38 & 69.1   & $\geq$0.556 \\  
\cline{2-17} 
& RANK             & 2 & 2 & 2 & 1 & 1 & 1 & 1 & 9 & 5 & 1 & 3 & 1 & 2 & 2 & 2 \\  
\bottomrule
		\end{tabular}}
\caption{Results of AnyBURL compared to current state of the art.}
	\label{tab:main}
\end{table*}

The test and validation set of FB15-237 have been filtered by removing triples that connect two entities which are already connected in the training set. According to the training set of FB15-237 we are sometimes right to say that $h(c,c')$ holds, if we already know that $b(c,c')$ or $b(c',c)$ holds. Contrary to this, the specific setup will punish such conclusions. 
We check prior to the prediction, whether the validation set connects entities not connected in the training set. If this is not the case, we block any prediction of a triple with two entities that are already connected. Note that this setting, which is always activated, has no impact on any other dataset, while it improves results for FB15-237 by $\approx$2\%. 

AnyBURL is not capable of learning rules with a head such as $h(X,X)$. This means that it cannot predict that an entity is related to itself via $h$. However, in some of the datasets (FB15 and FB15-237), a small subset in the training and test sets consist  such triples. 
To allow AnyBURL to learn meaningful rules in such a situation, we rewrite triples like $h(c,c)$ to $h(c,\mathit{self})$ by introducing a new constant $\mathit{self}$. Thus, AnyBURL can, for example, learn a \unaryex\ or \unarycon\ rule such as $h(X,\mathit{self}) \leftarrow b(X, A)$ or $h(X,\mathit{self}) \leftarrow b(X, c)$. After applying the rules, we convert a prediction of the form $h(c,\mathit{self})$ into $h(c,c)$. 

We have ranked all approaches for each combination of metric and dataset and present in the last row the rank that was achieved by applying the rules that AnyBURL learned after 10000 seconds. AnyBURL is in six cases on the first position, in six cases on the second position, and in the remaining three cases on position 3, 5 and 9. ComplEx performs also quite well on some datasets, however, it is not among the best models for WN18 and WN18RR. Moreover, AnyBURL performs in particular very good if we look at the hits@1 score. For each dataset, AnyBURL is the best or the second best system if we look only at the top-ranked candidate. Another remarkable result, is the capability of AnyBURL to learn in short time a rule set, that is already competitive compared to the other approaches. This can be concluded from the results achieved after 100 seconds. It is worth noting that unlike embedding models, AnyBURL does not require time-consuming hyperparameter tuning to achieve these numbers.

If we compare our results against the previous version of AnyBURL there seems to be only a small improvement. However, this improvement takes place in a range where it is hard to make better predictions. When manually analysing some of the given predictions, we realized that the predictions made by AnyBURL are sometimes right, even though they are not specified as facts in test, training or validation. These new correct facts, that are counted as wrong predictions, might share a common characteristic. An approach that looks into the validation set might be able to tune its hyperparameters to avoid these predictions. AnyBURL cannot capture such a regularity.


\section{Related Work}
\label{sec:related}


Most knowledge graph completion techniques are based on the concept of embeddings. There are also some approaches that try to combine embeddings and rules. An example is the system Ruge~\cite{ruge2018}, which learns rules, materializes these rules, and injects the materialized triples as new training examples with soft labels into the process of learning the embedding. The authors report results on FB15 which are worse than the results achieved by AnyBURL after 100 seconds. The benefits of combining rules and embeddings can only be understood, if we know first how far one can get with each method on its own. With our work, we show that rules on their own perform surprisingly well, which should not be neglected in further work on combining embeddings and rules.


Recently, reinforcement learning has been used for the task of query answering in~\cite{das2018go,lin2018multi,xiong2017deeppath}. These approaches have been applied to knowledge graph completion. Similar to AnyBURL, they provide explanations, however, they rely on vector representations and not on symbols. 
While these approaches use a reward strategy for paths that lead to answer nodes, AnyBURL uses reward strategies for path profiles that provide paths which result into rules. Even though in~\cite{das2018go,lin2018multi} FB15-237 and WN18RR are used, the results are based on a different evaluation procedure and/or a different test data split. Under this evaluation set up, the reinforced approaches perform as good or worse as ConvE, which is included in Table~\ref{tab:main} for comparison.

\section{Conclusion}
\label{sec:conclusion}

We introduced two extensions of our rule mining system AnyBURL. We explained and argued, based on our experimental results, that a rule-based solution to the knowledge graph completion problem should be based on Object Identity. As second contribution we introduced a reinforcement learning technique to guide the sampling process in order to use available computational resources in a reasonable way. Both extensions are implemented in the new version of AnyBURL available at \url{http://web.informatik.uni-mannheim.de/AnyBURL/}. We have evaluated this new version and compared the results against current state of the art embedding based techniques. The results show that most of these approaches cannot achieve the predictive quality of our approach, nor can they explain their predictions, which is a significant disadvantage compared to a symbolic approach.


\bibliographystyle{plain}
\bibliography{mybib}

\end{document}